\documentclass[runningheads]{llncs}

\usepackage{graphicx}
\usepackage[skip=2pt]{caption}
\usepackage{arabtex}
\usepackage{hyperref}
\usepackage{parskip}

\usepackage{times}
\usepackage{latexsym}

\usepackage[T1]{fontenc}

\usepackage{utf8}
\setcode{utf8}
\usepackage{microtype}
\usepackage{listings}
\usepackage{inconsolata}
\begin{document}

\title{ArabIcros: AI-Powered Arabic Crossword Puzzle Generation for Educational Applications}
\author{Kamyar Zeinalipour \and Mohamed Zaky Saad \and Marco Maggini \and Marco Gori \\
        DIISM, University of Siena \\ Via Roma 56, Siena, Italy\\
        \texttt{\{kamyar.zeinalipour2, marco.maggini, marco.gori\}@unisi.it}\\
        \texttt{m.zakyanwarzakymo@student.unisi.it}}

\authorrunning{K. Zeinalipour et al.}
\institute{Università degli Studi di Siena , Via Roma 56, Siena, Italy}

\maketitle

\begin{abstract}
This paper presents the first Arabic crossword puzzle generator driven by advanced AI technology. Leveraging cutting-edge large language models including GPT4, GPT3-Davinci, GPT3-Curie, GPT3-Babbage, GPT3-Ada, and BERT, the system generates distinctive and challenging clues. Based on a dataset comprising over 50,000 clue-answer pairs, the generator employs fine-tuning, few/zero-shot learning strategies, and rigorous quality-checking protocols to enforce the generation of high-quality clue-answer pairs. Importantly, educational crosswords contribute to enhancing memory, expanding vocabulary, and promoting problem-solving skills, thereby augmenting the learning experience through a fun and engaging approach, reshaping the landscape of traditional learning methods. The overall system can be exploited as a powerful educational tool that amalgamates AI and innovative learning techniques, heralding a transformative era for Arabic crossword puzzles and the intersection of technology and education.
\end{abstract}

\section{Introduction}\label{sec:Introduction}
Combining traditional puzzle constructs with educational components, pedagogical crosswords foster interactive learning experiences by integrating vocabulary, history, sciences, and other subjects. Intriguingly, they effectively strengthen students' vocabulary and spelling abilities due to the puzzles' requirement for accurate spelling \cite{orawiwatnakul2013crossword,dzulfikri2016application,bella2023improving}. These puzzles are particularly significant for language acquisition and learning specific technical terms \cite{nickerson1977crossword,sandiuc2020use,yuriev2016crossword}. Moreover, they enhance problem-solving, critical thinking skills, and memory retention, thereby making the learning process enjoyable and productive \cite{kaynak2023effect,dol2017gpbl,mueller2018testing,dzulfikri2016application,zirawaga2017gaming,bella2023improving,zamani2021use,yuriev2016crossword}.\\
Creating Arabic educational crosswords can be challenging due to the required wordplay expertise. However, with the help of innovations in natural language processing, Large Language Models (LLMs) are now able to generate high-quality Arabic crossword clues. LLMs are pre-trained on a mix of sources like books, academic articles, and web content and this wide spectrum of content enables them to create challenging and engaging crossword clues. This aids puzzle designers and improves the solver's experience, enabling even beginners to design personalized puzzles.\\
The results show that the proposed approach can be effectively employed to generate Arabic educational crossword puzzles, introducing an innovative system using LLMs to generate top-quality clues and answers. By inputting text passages or keywords, the system generates clue-answer pairs, based on techniques like fine-tuning and few-shot learning used for generation. We also present models to filter inappropriate clue-answer pairs for puzzle construction optimization, propose an advanced algorithm for designing Arabic educational crossword layouts, and provide a comprehensive dataset of curated Arabic clue-answer pairs. These advances simplify the creation of Arabic pedagogical crosswords and expand their potential for their broader exploitation.\\
This paper is structured as follows; section \ref{sec:relatedworks} explores relevant literature; section \ref{sec:dataset} discusses the collected Arabic dataset; section \ref{sec:methodology} outlines the research methodology, section \ref{sec:experiments} presents the findings, and, finally, section \ref{sec:conclusions} summarizes the overall outcomes.

\section{Related works}\label{sec:relatedworks}
The generation of crosswords represents a complex task that has been addressed by some research works. These studies have utilized a variety of tools, including traditional dictionaries and thesauri, or have engaged in the linguistic analysis of text content derived from the web.\\
Rigutini et al. \cite{rigutini2008fully,rigutini2012automatic} pioneered the first fully automated crossword creator system in 2008. The proposed system leverages natural language processing techniques to generate crossword clues by scraping related documents from the web, extracting relevant text segments, and using part-of-speech tagging, dependency parsing, and WordNet-based similarity measures. This approach produces clues based on specific ranking criteria.\\
An alternative methodology for crossword construction using natural language processing is documented in \cite{ranaivo2013automatic}. This approach consists of a four-stage process, which includes initial data retrieval of a targeted topic-specific text compilation, extraction of complete sentences, determination of the dependency syntactic structure of each sentence, and removal of words from stop lists. The extracted information undergoes a transformation into a graph representation for depth-first pre-order search. This framework integrates pre-processing, candidate identification, clue formation, and answer selection.\\
Esteche et al.'s study \cite{esteche2017automatic} delved into the creation of Spanish language crossword puzzles from news articles. The system is based on a twofold procedure: initially, pivotal terms are identified and their meanings are isolated from a trusted online dictionary. Subsequently, these definitions are employed as hints for the assembly of compelling crossword puzzles.\\
In a related study, Arora et al. \cite{arora2019automatic} discuss a software tool that uses NLP techniques to identify crucial keywords for creating crossword puzzles in various Indian languages. Their proposed framework, SEEKH, combines statistical and linguistic methods to highlight significant keywords useful for crossword creation.\\
Despite significant research, accurately generating comprehensive and unique clue-answer sets from linguistic corpora remains a challenge, particularly for the unique linguistic nuances of Arabic. To address these issues, we propose an innovative methodology using LLMs to create intricate educational clues. As a pioneering attempt, our technique successfully generates Arabic crossword puzzles, filling a gap unaddressed by previous methods. By generating intellectually stimulating and original crossword puzzles, this novel approach enhances learners' deep understanding of the subjects by providing comprehensive answers. Hence, the proposed work not only brings novelty to Arabic crossword generation but also offers a groundbreaking solution in the realm of educational tools.

\section{Dataset}\label{sec:dataset}
\begin{figure*}[h]
    \begin{center}
       \includegraphics[width = 1\textwidth]{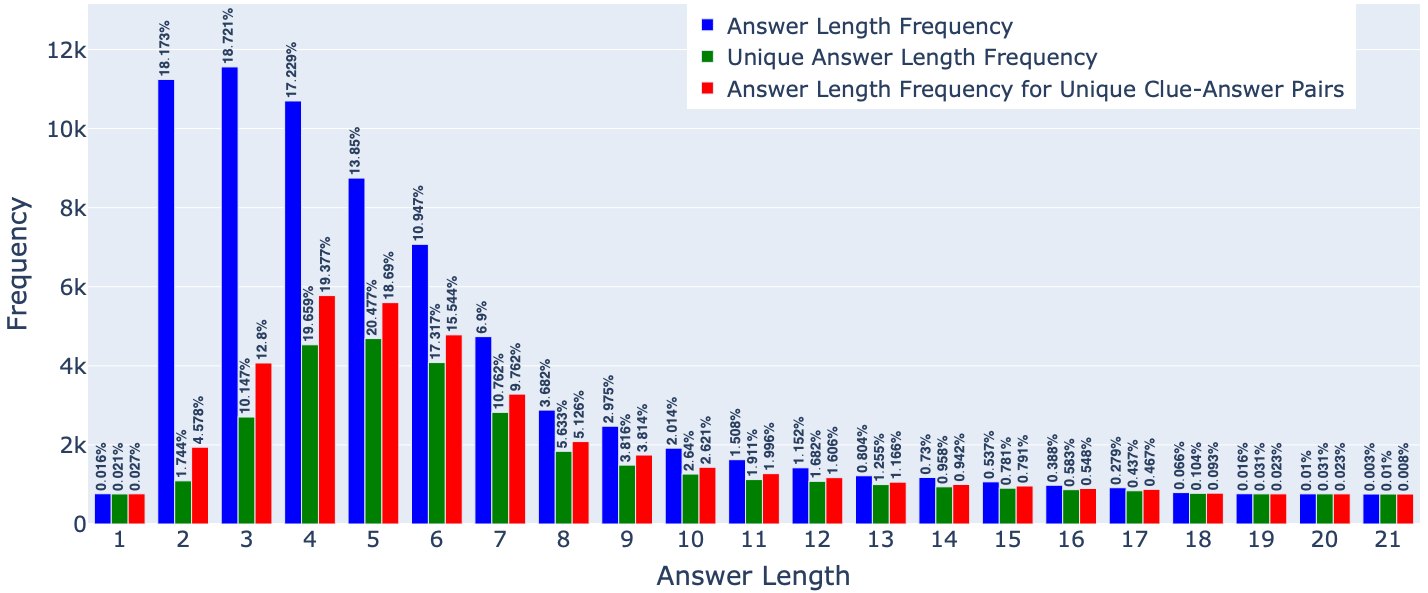} 
    \end{center}
    \caption{The introduced dataset entries are visually presented in terms of answer length distribution. The blue bars represent all the clue-answer pairs, while the green bars depict the frequency of unique answers. Additionally, the red bars indicate the frequency of unique answer-clue pairs.}
    \label{fig:1}
\end{figure*}

Given the scarcity of data for Arabic crossword puzzles, a clue-answer pair dataset was gathered manually. The dataset encompasses the period from 2020 to 2023.\\
During the initial stage of data collection, we pursued all accessible crossword puzzles, encompassing web-based games, journals, and magazines, ensuring that the training set comprised accurate clue-answer pairs sourced from original Arabic crossword puzzles. 
We had a collection of crossword images, and we needed to extract the text contained within these images to build a dataset for obtaining the text from these images. To accomplish this, we initially utilized optical character recognition (OCR) as a tool. However, it's important to note that the OCR process was predominantly supervised by humans who used it to facilitate the extraction. Additionally, human validation was employed to evaluate both spelling errors within the journals and the overall quality of the clue-answer pairs. This meticulous process resulted in a catalog of 57,706 entries from two different sources. One of them was the Al-Joumhouria Journal, from which we manually extracted 5,661 Clue and Answer pairs. The other source was the Al-Ghad Electronic Journal, where we utilized the OCR tool to assist in the extraction process. In the end, this yielded 25,908 unique pairs with answers varying in length from 1 to 21 characters, with the majority of the data falling within a specific answer's character length range from 2 to 9 (see Fig.~\ref{fig:1}).\\
The structure of the pairs is recurrent. For instance, some of the pairs are synonyms or antonym definitions,
that define the answer by means of one or more synonyms or antonyms. An example of this category includes "{\small\<موتي>}" with the answer "{\small\<حتفي>}". Some others were general information, such as for example "{\small\<دولة عربية>}" with the answer "{\small\<مصر>}". Another structure can be a word but the letters are not in order, as for example "{\small\<جميل مبعثرة>}" with the answer "{\small\<ل م ي ج>}". Finally, the definition can give the word and requires part of it for the answer, as for instance "{\small\<نصف نادر>}" with the answer "{\small\<در>}".\\
A meticulous pre-processing step was carried out on the data to refine it for fine-tuning. This involved the elimination of Arabic accents, redundant pairs, and markers suggesting a reversal in crosswords—an idiosyncrasy of Arabic. The aim of this study was to pave the way for further research by making this processed dataset \ref{dataset} publicly accessible, encouraging other scholars to contribute to this field.

\footnote{The dataset is available at \url{https://huggingface.co/datasets/Kamyar-zeinalipour/AR_CW} \label{dataset}}

\section{Methodology}\label{sec:methodology}
\begin{figure*}
    \begin{center}
       \includegraphics[width = 1\textwidth]{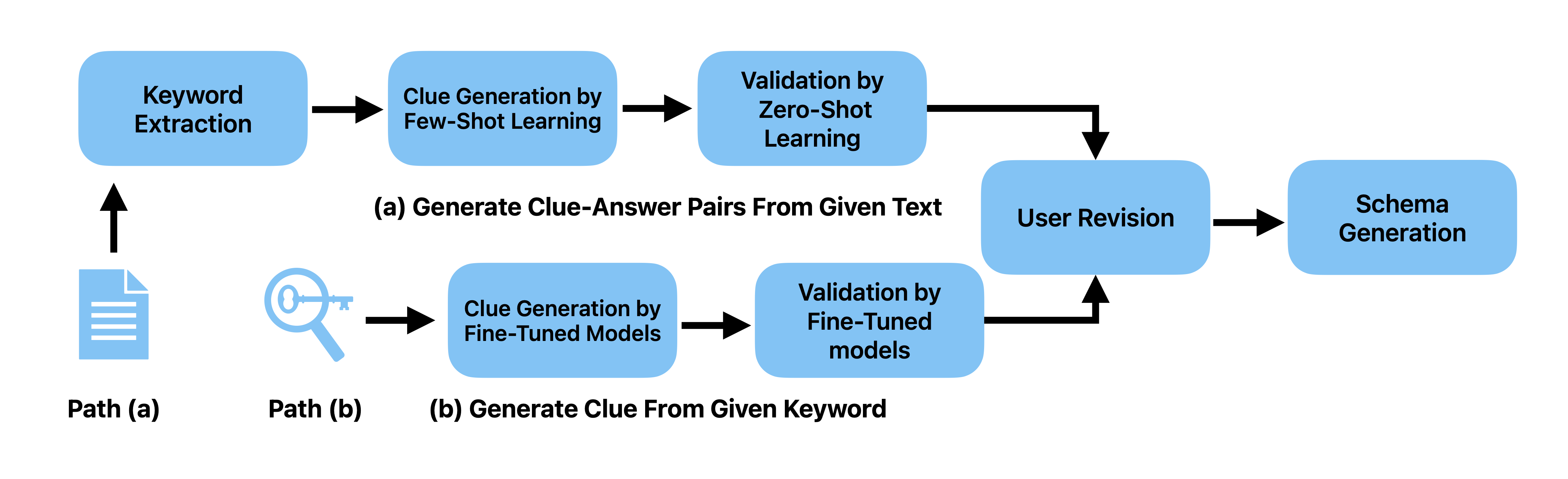} 
    \end{center}
    \caption{Overall system architecture. Path (a) Clue-answer generation from input text. Path (b) Clue generation from the given answers. } 
    \label{fig:2}
\end{figure*}

The proposed system includes several components, such as mechanisms to generate clue-answer pairs using user-provided text or keywords, and a crossword schema generator as depicted in Figure~\ref{fig:2}. Users can input any instructional text to extract relevant clue-answer pairs or insert a list of chosen keywords to generate clues. After combining both clue-generation methods, the quality of the generated pairs is evaluated using specific validation modules. Users can then review and select their preferred clue-answer sets, which are employed in the final step by a separate module for creating the crossword layout.
\subsection{Path (a): Generating clue-answer pairs from input text}
In our system, we employ zero-shot and few-shot learning to create clue-answer pairs. This process involves segmenting the text into paragraphs, keyword extraction, generating potential clues, and rigorously validating the resulting pairs. More details on these stages are provided later in the paper. Our experiments are based on the models GPT3.5-Turbo and GPT4 \cite{brown2020language} \cite{openai2023gpt4}. We use dynamic experimental approaches, including both customized English and Arabic prompts, to assess prompt language strategies' effectiveness across models.
\subsubsection{Keyword extraction}
\hfill \break
Our Few-Shot Learning Framework begins with prompt construction, involving the incorporation of extensive educational text that includes potential crossword keywords. These keywords, chosen to match possible answers from the provided text, enhance precision as the LLM is prompted with well-curated information. The process concludes by inputting the educational text and the tailored prompt to the LLM, enabling it to utilize its few-shot learning experiences to extract potential keywords from the input paragraph. This mechanism allows the LLM to extrapolate potential keywords effectively, resulting in a more comprehensive analysis.
\subsubsection{Generating crossword clues from the extracted keywords}
\hfill \break
In this stage, we harness the power of few-shot learning once more. By utilizing the keywords identified in the previous phase along with the input text, we generate relevant crossword clues. Additional information, including an example of valid paragraph, keywords, and clues, was also input into the LLM along with the target text and previously generated keywords that needed crossword clues. This strategy enabled the LLM to craft unique clues by leveraging the supplied text and initial keywords. This systematic approach significantly improves the precision and relevance of the generated crossword clues, ensuring each clue aligns with the context of the provided text and identified keywords.
\subsubsection{Path(a) Validation}
\label{sec:validation_1}
\hfill \break
To enhance the quality and appropriateness of our generated keyword-clue pairs, a method to exclude low-quality and inappropriate pairings is applied in several discrete stages. The first step utilized a filter system to eliminate answers containing more than three words, which are typically unsuitable for crossword puzzles. Our empirical research has shown that the LLM can occasionally produce clues by drawing upon its innate knowledge rather than relying solely on the provided text. Additionally, in instances where the generated clues did not effectively capture relevant keywords, we took steps to address this issue. To enhance the quality of our output and ensure the creation of appropriate clue-answer pairs, we employed a zero-shot learning approach, effectively filtering out undesired clues.
\subsection{Path (b): Generating clues based on provided answers} 
\hfill \break
There may be scenarios where we need to generate crossword clues using provided answers without a full-text context. To face this task, we deployed a holistic approach that started with fine-tuning different language models using the introduced Section \ref{sec:dataset}, each specifically designed for this task. We further enriched this scheme by using data from these fine-tuned models to create various classifiers. These classifiers aim to differentiate between high-quality generated clue-answer pairs and less suitable alternatives.
\subsubsection{Fine-tuning LLMs to generate clues from provided answers}
\label{sec:models}
\hfill \break
In the pursuit of crafting crossword clues from given answers and textual information, our research delved into the optimization of language models. This refinement process was informed by the dataset meticulously outlined in Section \ref{sec:dataset}. Our evaluation encompassed a spectrum of models, notably the robust Turbo 20B and the efficient Davinci 175B \cite{bongini2023gpt}, distinguished by their substantial 175 billion and 20 billion parameters, respectively. We also examined the 13 billion-parameter GPT3-Curie, recognized for its versatility,

This section encapsulates our methodical approach to model selection, emphasizing the diversity of parameters and architectures considered in our quest to enhance the generation of crossword clues. The subsequent analysis and results, detailed in the following sections, shed light on the efficacy and performance of these fine-tuned language models in the context of crossword clue generation.

\subsubsection{Path(b) Validation}
\hfill \break
The design of the overall system focuses on enhancing the overall quality of the generated clue-answer pairs. We incorporated a filtering process into the system pipeline to enhance the quality and usability of the generated pairs. Using the data obtained from the fine-tuned language models, we created a classifier capable of distinguishing between effective and unsuitable clues.\\
For this purpose, several models were fine-tuned, including GPT3-DaVinci with 175 billion parameters, GPT3-Curie with 13 billion parameters, GPT3-Babbage with 1.3 billion parameters, GPT3-Ada with 350 million parameters \cite{brown2020language}, and BERT-base-Arabic with 110 million parameters \cite{raffel2020exploring,safaya2020kuisail}. These models provided important insights into their respective capabilities and aided in validating the generated clues.\\
Our primary objective was to use these models with their varying parameter counts to comprehensively evaluate their effectiveness in filtering and validating the generated clues. This methodology aimed to ensure only high-quality and contextually relevant clues were retained, thereby improving the overall precision and functionality of our system.
\subsection{Schema Generator}
The algorithm for creating educational crossword puzzles follows a streamlined approach using input parameters such as the answer list, workspace dimensions, and termination criteria. Initially, a central answer is placed randomly followed by strategically adding surrounding answers. This cycle of adding and occasionally removing the recently added answers or entirely resetting is repeated until an optimal solution is obtained. The quality of the crossword is evaluated through a comprehensive scoring process. Each solution's merit is determined by the following scoring formula:
\begin{equation}
    \mathrm{Score} = (\mathrm{FW} + 0.5 \cdot \mathrm{LL}) \cdot \mathrm{FR} \cdot  \mathrm{LR} 
\end{equation}
The variables exploited in this formula correspond to the following metrics:
\begin{itemize}
    \item Filled Words ($\mathrm{FW}$): This represents the count of the added words, signaling the puzzle's completeness.
    \item Linked Letters ($\mathrm{LL}$): This counts the instances of letter-sharing between intersecting words, indicating the puzzle's coherence.
    \item Filled Ratio ($\mathrm{FR}$): This metric, calculated as the filled letters count divided by the area of the smallest covering rectangle, showcases the efficiency of the crossword's space utilization.
    \item Linked Letters Ratio ($\mathrm{LR}$): By dividing $\mathrm{LL}$ by the total letter count, $\mathrm{LR}$ highlights the extent of letter linkage and word-relations within the puzzle. 
\end{itemize}
These four criteria collectively contribute to the evaluation and selection of the optimal solution during the algorithm execution.\\
The algorithm makes use of a variety of stopping criteria to guide its decision-making and determine when to end the crossword construction. These criteria are as follows:
\begin{itemize}
    \item Minimum Number of Answers: The algorithm stops once it has added a preset minimum count of answers to the grid, ensuring an adequate crossword complexity.
    \item Minimum Filled Ratio Threshold: A certain threshold of the filled ratio, when met or surpassed, triggers the algorithm to stop, preventing the overabundance of empty spaces and maintaining appealing aesthetics.
    \item Grid Rebuilding Limit: The algorithm ceases to operate if the grid's reconstruction exceeds a set count, avoiding getting stuck in inefficient solutions and encouraging exploration of other possibilities.
    \item Maximum Time Duration: Upon reaching the allowed maximum time duration, the algorithm finishes, ensuring the process is time-efficient and the resources are optimally utilized.
\end{itemize}
This method allows the algorithm to identify the highest-scoring solution, enabling efficient production of high-quality crosswords given its input parameters. Furthermore, the algorithm can prioritize a list of "preferred answers," increasing their chances of inclusion, thereby ensuring that the crossword design aligns with specific objectives or preferences.
\section{Experiments}\label{sec:experiments}
In this section, we detail the empirical evaluation of the proposed system, focusing on individual elements and their roles within the overall framework.
\subsection{Experimental Evaluation: Path (a)}
\begin{figure*}[h]
    \begin{center}
       \includegraphics[width = 1\textwidth]{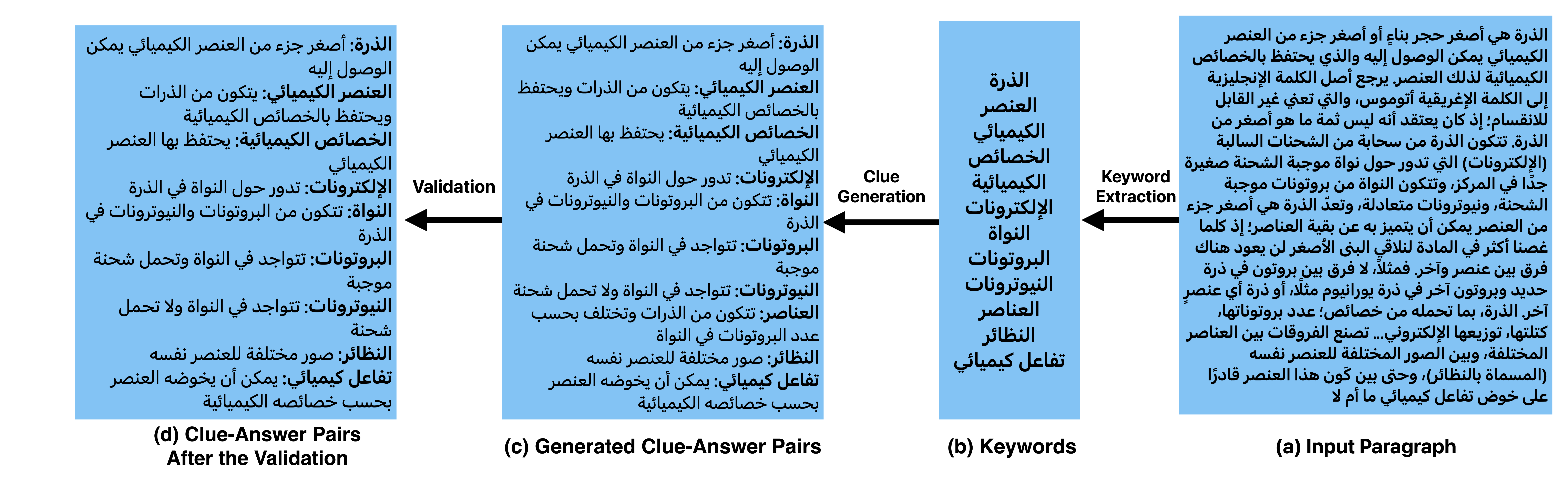} 
    \end{center}
    \caption{A comprehensive collection of clue-answer pairs generated by the introduced system from a given text, providing illustrative examples.}
    \label{fig:3}
\end{figure*}

This paper's experimental dataset aims to rigorously assess our system's output quality in relation to various language prompts. We conducted an in-depth investigation using two prompt types, categorized as English and Arabic. Two different models, GPT4 and GPT3.5 Turbo, were used for evaluation. The comprehensive list of prompts can be found within the paper's Appendix \ref{sec:appendix_b}. This provides comprehensive evaluations of linguistic aspects, leading to robust, multifaceted findings. The system underwent thorough evaluation using 100 educational selected Wikipedia paragraphs to examine performance in different language contexts. Performance markers were established based on empirical evidence. Evaluation guidelines, created under expert supervision, ensured robust results. Detailed criteria for evaluation are in Appendix \ref{sec:appendix}, and cumulative findings are presented in Table \ref{tab:prompt}.
\begin{table*}[htbp]
\caption{Assessment outcomes of the clue-answer pairs generated from the provided Text.}
\begin{center}
\begin{tabular}{llll}
\hline
\textit{Model} &\textit{System Part} & \textit{English Prompt} & \textit{Arabic Prompt} \\
\hline
GPT4 & Keyword Extractor &   95.05 \%    &   94.32\%  \\
&Clues Generator  &   94.62 \%    &   93.23 \%  \\
&Validator  &   87.76 \%    &   84.01 \%  \\
&Total performance   &   78.95 \%    &   74.6 \%  \\
\hline
GPT3.5-Turbo &Keyword Extractor &   92 \%    &   97.38\%  \\
&Clues Generator &   55.33 \%    &   37.78 \%  \\
&Validator    &   89.04 \%    &   89.32 \% \\
&Total performance   &   46.68 \%    &   68.83 \%  \\

\hline
\end{tabular}
\label{tab:prompt}
\end{center}
\end{table*}
GPT4 and GPT3.5-Turbo models performed impressively in English prompts, achieving keyword extraction accuracy of 95.05\% and 92\% respectively. They similarly excelled in Arabic prompts with accuracies of 94.32\% and 97.38\%.\\
In clue generation, these models demonstrated their value in retrieving meaningful information. In English prompts, GPT4 and GPT3.5-Turbo reached accuracies of 94.62\% and 55.33\%, respectively, while GPT4 and GPT3. marking respective accuracies of 93.23\% and 37.78\% in Arabic prompts.\\
The evaluation of clue-answer pairs yielded satisfactory results. In English, the GPT4 and GPT3.5-Turbo  models exhibited accuracies of 87.76\% and 89.04\% and maintained substantial accuracy of 84.01\% and 89.32\% in Arabic prompts.\\
In the final evaluation, which included system-wide validation and acceptability of potentially generated clues and answers, both models upheld their performance. it means we analyze the clue-answer pairs that align with the validation part of the system, and then culminate in the calculation of the proportion of generated clues and answers that successfully pass the criteria established through human oversight which is the total performance of the model. It was overall 78.95\% and 74.6\% for the GPT4 model for English and Arabic prompts, respectively, while the GPT3.5-Turbo model had a total performance of 46.68\% and 68.83\% for English and Arabic prompts respectively.\\
Figure \ref{fig:3} provides a practical illustration of this system component's functionality. It sequentially depicts the transformation from initial text to final crossword clue-answer pairs, demonstrating input paragraphs (a), keyword extraction (b), clue generation (c), and clue-answer pair validation (d). This visual representation clarified the system's operational process, elucidating its capability to turn text into precise crossword clues and their corresponding answers. Comprehensive translations for the content depicted in Figure \ref{fig:3} can be found in the paper's Appendix \ref{sec:appendix_c}.
\subsection{Experimental Evaluation: Path (b)}
This section details experimental tests on clue generation and validation from keywords using three distinct models, GPT3-DaVinci, GPT3.5-Turbo, and GPT3-Curie. These were designed and optimized based on concepts discussed in Section \ref{sec:models}, with a specific emphasis on forming clues from identified keywords.\\
In the preparation phase, a subset of the dataset discussed in Section \ref{sec:dataset}, specifically 25,908 unique clue-answer pairs, was selected. Afterwards, each refined model produced 2,000 clues which were evaluated using human judgement based on the criteria presented in Appendix \ref{sec:appendix}.\\
In conclusion of our evaluation, Table \ref{tab:pathb} presents the results, highlighting the performances of GPT3-DaVinci, GPT3.5-Turbo, and GPT3-Curie. These models successfully generated satisfactory clues 41.9\%, 81\%, and 21.35\% of the time, respectively. Observations indicate that GPT3.5-Turbo significantly outperforms the other models in the task of clue generation from the given keywords.
\begin{table}[htbp]
\caption{Assessment outcomes of the clues generated from the provided keyword.}
    \begin{center}
        \begin{tabular}{ll}
        \hline
        \textit{Model}&  \textit{\% of acceptable  clues}  \\
        \hline
        GPT3-DaVinci& 41.9   \\
        GPT3-Curie& 21.35  \\
        GPT3.5-Turbo & 81 \\ 
        \hline
        \end{tabular}
        \label{tab:pathb}
    \end{center}
\end{table}
For a thorough assessment of the generated clues, a detailed review identifying acceptable and unacceptable cases was undertaken. Each clue-answer pair was carefully examined and categorized, Tables \ref{tab:valexample1} and \ref{tab:valexample2} present illustrative clues generated by distinct fine-tuned models. Table \ref{tab:valexample1} demonstrates instances of well-constructed clues, while Table \ref{tab:valexample2} highlights cases of unacceptable clue generation. Detailed translations for these clues can be located in the Appendix \ref{sec:appendix_c}. This meticulous evaluation facilitated performance analysis of the algorithm, notably its ability to generate captivating crossword puzzles.
\begin{table}[h]
\caption{Acceptable clues from given keywords using various models.}
\begin{center}
\begin{tabular}{rcl}
\hline
\textit{Model} & \textit{Clue-Answer pair}  \\
\hline
 GPT3-DaVinci & {\small\< نجوم : في السماء ليلا> } \\

 GPT3-Curie & {\small\<كروم : من المعادن> } \\

 GPT3.5-Turbo & {\small\<قوة : قدرة >  }\\
\hline

\end{tabular}
\label{tab:valexample1}
\end{center}
\end{table}

\begin{table}[h]
\caption{Unacceptable clues from given keywords using various models.}
\begin{center}
\begin{tabular}{rcl}
\hline
 \textit{Model} & \textit{Clue-Answer pair}  \\
\hline

 GPT3-DaVinci & {\small\<زرافة :  من الحشرات> }\\
 GPT3-Curie & {\small\<مثلث :   مثنى مثلث >} \\
 GPT3.5-Turbo & {\small\<عمة : اخت والد او والدة>} \\
\hline
\end{tabular}
\label{tab:valexample2}
\end{center}
\end{table}
Several classifiers were developed in this study. Coupled with various language models, they enabled the distinction between suitable and unsuitable clue-answer pairings. The results from the evaluation of the test set are shown in Table \ref{tab3}.\\
The process utilized a dataset of 6,000 human-evaluated instances from previous steps to build several classifiers. The dataset was divided, with 80\% used for training, and the remaining 20\% for testing classifier performance. The analysis revealed that the dataset consists of 52\% acceptable clues and 48\% unacceptable ones.
\begin{table*}[htbp]
\caption{Classifier performance on distinguishing acceptable Clue-Answer pairs}
\begin{center}
\begin{tabular}{lllll}

\hline
\textit{Model}&  \textit{accuracy \%} & \textit{precision \%}& \textit{recall \%}& \textit{F1 Score}  \\
\hline
GPT3-Dvinci & 85.74  & 83.39    & 85.26  & 0.8431\\
GPT3-Curie  &81.29  & 78.86   & 79.89  & 0.7937\\
GPT3-Babbage& 78.69  & 75.17    &78.54  & 0.7682\\
GPT3-Ada    & 79.19  & 77.48    & 75.75  & 0.7660\\
Bert-base-Arabic & 71.42  & 67.91    & 70.04  & 0.6896\\
\hline
\end{tabular}
\label{tab3}
\end{center}
\end{table*}
The system's effectiveness was gauged through the accuracy of four distinct classifiers - GPT3-DaVinci, GPT3-Curie, GPT3-Babbage, GPT3-Ada, and Bert in discerning between satisfactory and unsatisfactory clues. Notably, GPT3-DaVinci topped the list with an exceptional 85.74\% accuracy, followed by GPT3-Curie at 81.29\%. GPT3-Babbage showed decent results with 78.69\% accuracy, while GPT3-Ada and Bert had fair performances with 79.19\% and 71.42\% accuracy, respectively. These results underscore the commendable performance of these classifiers in identifying agreeable clues.
\subsection{Schema Generation}
\begin{figure*}
    \begin{center}
       \includegraphics[width = 1\textwidth]{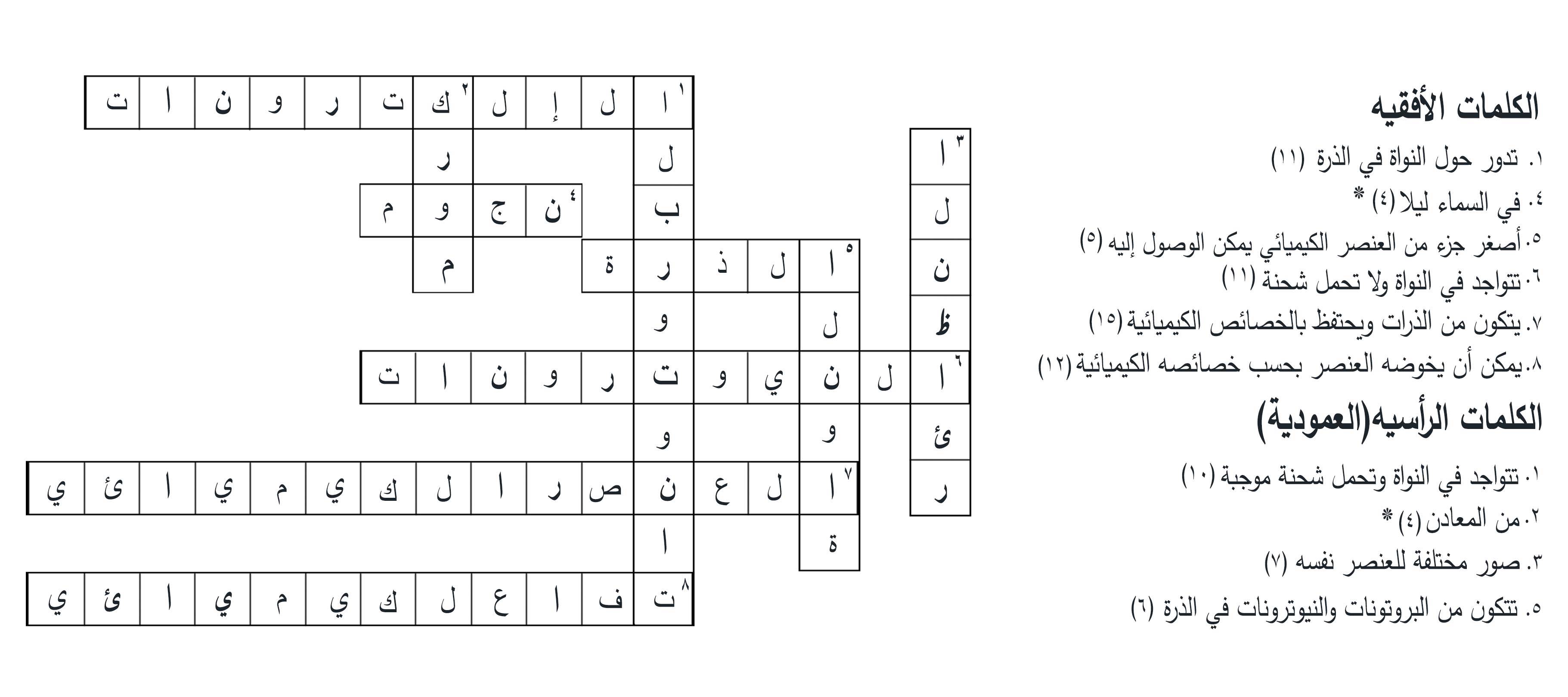} 
    \end{center}
    \caption{An illustrative Arabic educational crossword generated through the proposed system.}
    \label{fig:4}
\end{figure*}

Our algorithm for schema generation envisages a spectrum of educational crosswords utilizing a group of generated clue-answer pairs. Illustrated in Figure \ref{fig:4} is a comprehensive Arabic educational crossword about physics, crafted by the proposed system. The clue-answer pairs are procured either from a text (path (a), refer to Figure \ref{fig:3}) or directly produced from a keyword (path (b), denoted by examples marked with a $\star$), as observed in Table \ref{tab:valexample1}.

\section{Conclusions}\label{sec:conclusions}
The work featured in this paper focuses on multiple innovative offerings, among which is the introduction of a comprehensive dataset for Arabic clue-answer pairs. In addition to this, we have also formulated a ground-breaking method employing large language models that generate educational Arabic crossword puzzles influenced by the provided texts or given keywords.\\
To uphold stringent quality standards in our methodology, our approach integrates human oversight in conjunction with specific guidelines (see Appendix \ref{sec:appendix}). In the process of generating clue-answer pairs from textual data, we conducted experiments using two distinct models: GPT-4 and GPT3.5-Turbo, while employing prompts in both English and Arabic languages.  We conducted various types of evaluations considering different parts of the system and overall performance:
\begin{itemize}
    \item Keyword Extraction: Notably, when paired with Arabic prompts, GPT3.5-Turbo exhibited exceptional performance, successfully generating high-quality keywords with an impressive accuracy rate of 97.38\%.
    \item Crossword Clue Generation: GPT4, when prompted in English, consistently produced relevant and well-suited crossword clues, achieving a commendable success rate of 94.62\%.
    \item Validation Component: Within our system, the validation step was a critical component. GPT3.5-Turbo, when prompted in Arabic, demonstrated superior performance in this role, boasting an impressive validation accuracy rate of 89.32\%.
    \item Total Performance: GPT4 displayed remarkable proficiency in this role, surpassing expectations with an impressive validation accuracy rate of 78.95\% when prompted in English.
\end{itemize}
In our quest to generate clues from provided keywords, we engaged in the fine-tuning process using a curated dataset (refer to Section \ref{sec:dataset}). We fine-tuned three distinct models, namely GPT3-DaVinci, GPT3.5-Turbo, and GPT3-Curie. We rigorously tested the performance of each model by generating clues for a carefully chosen set of 2000 educational-related keywords. Notably, the fine-tuned GPT3.5-Turbo outperformed the others, consistently producing high-quality clues with a remarkable success rate of 81\%.\\
Utilizing the data generated through the evaluation of fine-tuned models, we construct classifiers to distinguish between acceptable and non-acceptable clues for a specified keyword. The most effective model in this task was GPT3-Davinci, achieving an impressive accuracy rate of 85.74\%.\\
Our process to produce educational crossword layouts is both efficient and diverse. We hope that these findings will enrich the learning process and foster interactive learning. The developed system can be integrated into current teaching methods to enhance educational practices. As a future course of action, we plan on venturing into the development of more advanced models for more direct clue and answer pair generation and examine specialized models for different clue types. We also intend to implement this system in actual classrooms and evaluate its impact. Our goal is to revolutionize the creation of educational crossword puzzles and usher in an era of unique teaching practices.
\section*{Acknowledgments}

The funding for this paper was provided by the TAILOR project and the Humane-AI-Net projects, both supported by the EU Horizon 2020 research and innovation program under GA No 952215 and No 952026, respectively.
\newpage
\bibliographystyle{splncs04}

\bibliography{custom}

\begin{thebibliography}{10}
\providecommand{\url}[1]{\texttt{#1}}
\providecommand{\urlprefix}{URL }
\providecommand{\doi}[1]{https://doi.org/#1}

\bibitem{arora2019automatic}
Arora, B., Kumar, N.: Automatic keyword extraction and crossword generation tool for indian languages: Seekh. In: 2019 IEEE Tenth International Conference on Technology for Education (T4E). pp. 272--273. IEEE (2019)

\bibitem{bella2023improving}
Bella, Y.D., Rahayu, E.M.: The improving of the student’s vocabulary achievement through crossword game in the new normal era. Edunesia: Jurnal Ilmiah Pendidikan  \textbf{4}(2),  830--842 (2023)

\bibitem{bongini2023gpt}
Bongini, P., Becattini, F., Del~Bimbo, A.: Is gpt-3 all you need for visual question answering in cultural heritage? In: Computer Vision--ECCV 2022 Workshops: Tel Aviv, Israel, October 23--27, 2022, Proceedings, Part I. pp. 268--281. Springer (2023)

\bibitem{brown2020language}
Brown, T., Mann, B., Ryder, N., Subbiah, M., Kaplan, J.D., Dhariwal, P., Neelakantan, A., Shyam, P., Sastry, G., Askell, A., et~al.: Language models are few-shot learners. Advances in neural information processing systems  \textbf{33},  1877--1901 (2020)

\bibitem{dol2017gpbl}
Dol, S.M.: Gpbl: An effective way to improve critical thinking and problem solving skills in engineering education. J Engin Educ Trans  \textbf{30}(3),  103--13 (2017)

\bibitem{dzulfikri2016application}
Dzulfikri, D.: Application-based crossword puzzles: Players’ perception and vocabulary retention. Studies in English Language and Education  \textbf{3}(2),  122--133 (2016)

\bibitem{esteche2017automatic}
Esteche, J., Romero, R., Chiruzzo, L., Ros{\'a}, A.: Automatic definition extraction and crossword generation from spanish news text. CLEI Electronic Journal  \textbf{20}(2) (2017)

\bibitem{kaynak2023effect}
Kaynak, S., Erg{\"u}n, S., Karada{\c{s}}, A.: The effect of crossword puzzle activity used in distance education on nursing students’ problem-solving and clinical decision-making skills: A comparative study. Nurse Education in Practice  \textbf{69},  103618 (2023)

\bibitem{mueller2018testing}
Mueller, S.T., Veinott, E.S.: Testing the effectiveness of crossword games on immediate and delayed memory for scientific vocabulary and concepts. In: CogSci (2018)

\bibitem{nickerson1977crossword}
Nickerson, R.: Crossword puzzles and lexical memory. In: Attention and performance VI, pp. 699--718. Routledge (1977)

\bibitem{openai2023gpt4}
OpenAI: Gpt-4 technical report (2023)

\bibitem{orawiwatnakul2013crossword}
Orawiwatnakul, W.: Crossword puzzles as a learning tool for vocabulary development. Electronic Journal of Research in Education Psychology  \textbf{11}(30),  413--428 (2013)

\bibitem{raffel2020exploring}
Raffel, C., Shazeer, N., Roberts, A., Lee, K., Narang, S., Matena, M., Zhou, Y., Li, W., Liu, P.J.: Exploring the limits of transfer learning with a unified text-to-text transformer. The Journal of Machine Learning Research  \textbf{21}(1),  5485--5551 (2020)

\bibitem{ranaivo2013automatic}
Ranaivo-Malan{\c{c}}on, B., Lim, T., Minoi, J.L., Jupit, A.J.R.: Automatic generation of fill-in clues and answers from raw texts for crosswords. In: 2013 8th International Conference on Information Technology in Asia (CITA). pp.~1--5. IEEE (2013)

\bibitem{rigutini2008fully}
Rigutini, L., Diligenti, M., Maggini, M., Gori, M.: A fully automatic crossword generator. In: 2008 Seventh International Conference on Machine Learning and Applications. pp. 362--367. IEEE (2008)

\bibitem{rigutini2012automatic}
Rigutini, L., Diligenti, M., Maggini, M., Gori, M.: Automatic generation of crossword puzzles. International Journal on Artificial Intelligence Tools  \textbf{21}(03),  1250014 (2012)

\bibitem{safaya2020kuisail}
Safaya, A., Abdullatif, M., Yuret, D.: Kuisail at semeval-2020 task 12: Bert-cnn for offensive speech identification in social media. arXiv preprint arXiv:2007.13184  (2020)

\bibitem{sandiuc2020use}
Sandiuc, C., Balagiu, A.: The use of crossword puzzles as a strategy to teach maritime english vocabulary. Scientific Bulletin" Mircea cel Batran" Naval Academy  \textbf{23}(1),  236A--242 (2020)

\bibitem{yuriev2016crossword}
Yuriev, E., Capuano, B., Short, J.L.: Crossword puzzles for chemistry education: learning goals beyond vocabulary. Chemistry education research and practice  \textbf{17}(3),  532--554 (2016)

\bibitem{zamani2021use}
Zamani, P., Haghighi, S.B., Ravanbakhsh, M.: The use of crossword puzzles as an educational tool. Journal of Advances in Medical Education \& Professionalism  \textbf{9}(2), ~102 (2021)

\bibitem{zirawaga2017gaming}
Zirawaga, V.S., Olusanya, A.I., Maduku, T.: Gaming in education: Using games as a support tool to teach history. Journal of Education and Practice  \textbf{8}(15),  55--64 (2017)

\end{thebibliography}
\appendix

\section{Appendix}
\label{sec:appendix}
This study entailed developing a classifier to distinguish optimal and sub-optimal crossword clue-answer pairs. Crossword puzzles necessitate linguistic acumen, innovation, and adherence to construction guidelines for quality clues and answers. Such a classifier auto-evaluates the clue-answer quality, aiding puzzle designers and improving puzzle-solving experiences. This provides insight into key aspects of language and puzzle architecture.\\
The development of a robust framework for determining acceptable and unacceptable crossword clue-answer pairs is crucial to the effectiveness of a classifier. This provides the groundwork upon which our classifier can effectively discriminate between high-quality clues and ill-fit ones. Rigorous adherence to these guidelines facilitates accuracy in quality evaluation by the classifier and ultimately enhances the appeal and satisfaction derived from crossword puzzles.\\
Let us now probe into the salient features of the guideline for assessing crossword clue-answer quality:
\begin{itemize}
    \item Coherence and Relevance: An ideal pair of clues and answers should display an evident and significant association between the two. The clue should offer adequate context or prompts that guide solvers toward the desired solution. The answer should be linear to the clue and sound logical within the subject matter or theme of the given puzzle.
    \item Wordplay and Creativity: A finely constructed crossword clue frequently employs wordplay, ingenious nuances, or concealed connotations that provoke and fascinate solvers. Seek clues that necessitate unconventional thinking, dual meanings, or linguistic resourcefulness. An effective clue-answer duo will enthrall the solvers, enhancing the puzzle's intrigue and pleasure.
    \item  Unambiguity and Specificity: Clues should be unequivocal and clear-cut, presenting solvers with a distinct and exact solution. Refrain from clues that allow for multiple interpretations or result in various potential answers. The aim is to propose a single accurate answer that correlates directly with the intended meaning of the clue.
    \item Linguistics and Grammar: Both the clue and the answer should conform to correct grammar, syntax, and language norms. It's essential to verify that the language utilized in the clue-answer duo is grammatically accurate, coherent, and appropriate for a crossword puzzle.
    \item Universal Knowledge and Equity: Clues should be based on general knowledge or facts that a wide spectrum of solvers would reasonably be anticipated to understand. Refrain from using excessively obscure or specialized references, which only a small subset of solvers would recognize. An optimal clue-answer match should maintain a balance between challenge and fairness, accommodating a varied assortment of puzzle aficionados.
\end{itemize}
Adhering to these guidelines, we can construct a dataset capable of building a dependable classifier to differentiate between well-formulated crossword clue-answer pairs and those that are nonsensical or inappropriate. This classifier holds the potential to transform the process of creating, evaluating, and solving crossword puzzles. It offers crucial insights into the art of crafting puzzles that are both engaging and intellectually challenging.

\section{Appendix}
\label{sec:appendix_b}
The following prompts were employed for (Keyword Generation, Clue Generation, and Clue Verification) in both the Arabic and English versions:

\textbf{English Keyword Extraction Prompt: }

"Objective: Your task is to extract 
keywords (maximum 2 words) from a given 
text to create short crossword definitions.
Please follow these steps to achieve the 
objective:

Keyword extraction: Extract the most 
important keywords from the text.

Validate keywords: Check if the
keywords are well explained 
in the given text.

Final keywords: Remove all the keywords 
that are not well-defined in the text,
based on the previous step.

Text: \{text\}

Here is an example Text:

{\small
\begin{arabtext}
الفقرة: "الأسد حيوان من الثدييات من فصيلة السنوريات وأحد السنوريات الأربعة الكبيرة المنتمية لجنس النمور ، وهو يُعد ثاني أكبر السنوريات في العالم بعد الببر، حيث تفوق كتلة الذكور الكبيرة منه 250 كيلوغراما (550 رطلًا). تعيش معظم الأسود البرية المتبقية اليوم في إفريقيا جنوب الصحراء الكبرى، ولا تزال جمهرة واحدة صغيرة مهددة بالانقراض تعيش في آسيا بولاية غوجرات في شمال غربي الهند. كان موطن الأسود شاسعًا جدًا في السابق، حيث كانت تتواجد في شمال إفريقيا، الشرق الأوسط، وآسيا الغربية، حيث انقرضت منذ بضعة قرون فقط. وحتى بداية العصر الحديث (الهولوسين، منذ حوالي 10,000 سنة)، كانت الأسود تُعتبر أكثر ثدييات اليابسة الكبرى انتشارا بعد الإنسان، حيث كانت توجد في معظم أنحاء إفريقيا، الكثير من أنحاء أوراسيا من أوروبا الغربية وصولا إلى الهند، وفي الأمريكيتين، من يوكون حتى البيرو."
\end{arabtext}}

Below are the legitimate keywords extracted from the provided text:

{\small
\begin{arabtext}
الكلمات المفتاحية: الأسد, الثدييات, فصيلة السنوريات, الأسود البرية, إفريقيا, الهند, شمال إفريقيا, الشرق الأوسط, آسيا الغربية, انتشار, الإنسان , نمور , ذكور     
\end{arabtext}}

Use the following output format:

Keywords: <Final keywords>"
\newpage

\textbf{English Clue Generation Prompt: }

"Your objective is to create short crossword clues for a list of keywords based on the given text:

Keywords: \{keywords\}
Text: \{text\}

Follow these steps to achieve the task:

Identify the part of the text that contains information about each provided keyword.

Generate short Arabic crossword clues (maximum 4 words) for all the keywords, using just the information from the text.

Here is an example Text:

        {\small
        \begin{arabtext}
         الفقرة: أسد حيوان من الثدييات من فصيلة السنوريات وأحد السنوريات الأربعة الكبيرة المنتمية لجنس النمور. وهو يُعد ثاني أكبر السنوريات في العالم بعد الببر، حيث تفوق كتلة الذكور الكبيرة منه 250 كيلوغراما (550 رطلًا). تعيش معظم الأسود البرية المتبقية اليوم في إفريقيا جنوب الصحراء الكبرى، ولا تزال جمهرة واحدة صغيرة مهددة بالانقراض تعيش في آسيا بولاية غوجرات في شمال غربي الهند. كان موطن الأسود شاسعًا جدًا في السابق، حيث كانت تتواجد في شمال إفريقيا، الشرق الأوسط، وآسيا الغربية، حيث انقرضت منذ بضعة قرون فقط. وحتى بداية العصر الحديث (الهولوسين، منذ حوالي 10,000 سنة)، كانت الأسود تُعتبر أكثر ثدييات اليابسة الكبرى انتشارا بعد الإنسان، حيث كانت توجد في معظم أنحاء إفريقيا، الكثير من أنحاء أوراسيا من أوروبا الغربية وصولا إلى الهند، وفي الأمريكيتين، من يوكون حتى البيرو.
        \end{arabtext}}

Below is a list of valid keywords for the provided text:
\newline  
    {\small
    \begin{arabtext}
     الكلمات المفتاحية: أسد, حيوان, ثدييات, سنوريات, سنوريات الأربعة, جنس النمور, ببر, الزكور الكبيرة, إفريقيا, صحراء الكبرى, أمريكيتين
    \end{arabtext}}

Here is a compilation of valid clue-answer pairs corresponding to the provided keywords and text:

Keyword: {\small\<أسد>}\\
Clue: {\small\<حيوان ثديي من السنوريات>}\\
\newline  
Keyword: {\small\<حيوان>}\\
Clue: {\small\<ينتمي لفصيلة السنوريات>}\\
\newline  
Keyword: {\small\<ثدييات>}\\
Clue: {\small\<نوع من الحيوانات>}\\
\newline  
Keyword: {\small\<سنوريات>}\\
Clue: {\small\<تشمل الأسد>}\\
\newline  
Keyword: {\small\<سنوريات الأربعة>}\\
Clue: {\small\<مجموعة من السنوريات الكبيرة>}\\
\newline  
Keyword: {\small\<جنس النمور>}\\
Clue: {\small\<يعتبر الأسد منه>}\\
\newline  
Keyword: {\small\<ببر>}\\
Clue: {\small\<السنورية الأكبر في العالم>}\\
\newline  
Keyword:{\small\<الزكور الكبير>}\\
Clue: {\small\<تتجاوز وزنها 250 كيلوغرام>}\\
\newline  
Keyword: {\small\<إفريقيا>}\\
Clue: {\small\<مكان عيش معظم الأسود البرية>}\\
\newline  
Keyword: {\small\<صحراء الكبرى>}\\
Clue: {\small\<تقع إلى جنوب إفريقيا>}\\
\newline  
Keyword: {\small\<أمريكيتين>}\\
Clue: {\small\<تتواجد الأسود فيهما>}\\
\newline  
Use the following format:\\
\newline  
Keyword: <Keyword>\\
\newline  
Clue: <Crossword Clue>\\
\newline  
\textbf{English Prompt for Hallucination Verification: }\\
\newline
"Please assess the quality of the
crossword clues based on the
given text.\\
\newline
Text: \{text\}\\
\newline
Clues: \{clues\}\\
\newline
To accomplish this task, follow these steps:\\
\newline
Check Clue in the text: Verify
Whether the content of each clue 
is present in the text.\\ 
\newline
If a content clue is found in the text, 
print True; otherwise, print False.\\
\newline
Use the following format for each clue:\\
\newline
Check Clue in the text: \\
\newline
<Check Clue in the text>"\\
\newline
\textbf{Arabic Keyword Generation Prompt: }\\
\newline
{
\begin{arabtext}
\small
       الهدف: استخراج كلمات مفتاحية (تتكون من كلمتين على الأكثر) من الفقرة التالية لإستخدام هذه الكلمات المفتاحية لإنشاء تعريفات قصيرة من اجل لعبة الكلمات المتاقطعة
        تتأكد من استخراج اهم الكلمات المفتاحية من الفقرة ثم قم بعمل فحص لهذه الكلمات المتقاطعة اذا كان تم شرحها بشكل جيد و واضح في الفقرة واذا لم تجد شرح وافي لكلمة من الكلمات المفتاحية فقم بالتخلص منها
        
        الفقرة: \end{arabtext}}\\ \{text\}

{\small
\begin{arabtext}
    
        مثال للمطلوب: 
        هذه الفقرة التي قمت بإستخراج منها الكلمات المفتاحية
        الفقرة: "الأسد حيوان من الثدييات من فصيلة السنوريات وأحد السنوريات الأربعة الكبيرة المنتمية لجنس النمور ، وهو يُعد ثاني أكبر السنوريات في العالم بعد الببر، حيث تفوق كتلة الذكور الكبيرة منه 250 كيلوغراما (550 رطلًا). تعيش معظم الأسود البرية المتبقية اليوم في إفريقيا جنوب الصحراء الكبرى، ولا تزال جمهرة واحدة صغيرة مهددة بالانقراض تعيش في آسيا بولاية غوجرات في شمال غربي الهند. كان موطن الأسود شاسعًا جدًا في السابق، حيث كانت تتواجد في شمال إفريقيا، الشرق الأوسط، وآسيا الغربية، حيث انقرضت منذ بضعة قرون فقط. وحتى بداية العصر الحديث (الهولوسين، منذ حوالي 10,000 سنة)، كانت الأسود تُعتبر أكثر ثدييات اليابسة الكبرى انتشارا بعد الإنسان، حيث كانت توجد في معظم أنحاء إفريقيا، الكثير من أنحاء أوراسيا من أوروبا الغربية وصولا إلى الهند، وفي الأمريكيتين، من يوكون حتى البيرو."
        
        الكلمات المفتاحية التي تم إستخراجها كالأتي
        الكلمات المفتاحية: الأسد, الثدييات, فصيلة السنوريات, الأسود البرية, إفريقيا, الهند, شمال إفريقيا, الشرق الأوسط, آسيا الغربية, انتشار, الإنسان , نمور , ذكور

       شكل النتيجة النهائية: 
       'الكلمات المفتاحية'

\end{arabtext}}

\textbf{Arabic Clue Generation Prompt: }

{\small
\begin{arabtext}

    هدفك هو إنشاء الغاز قصيرة للعبة الكلمات المتقاطعة مناسبة للكلمات المفتاحية الأتية استنادا الى الفقره 
        بحيث ان يكون كل كلمة مفتاحية يوجد لها اللغز خاص بها
        سأقوم بتزويدك بمثال بعد طريقة اتمام المهمة\\
        
        الكلمات المفتاحية: -الكلمات المفتاحية- \\
        الفقرة: -الفقرة-\\
        
         استخدم هذه الطريقة لإتمام المهمة: \\
         
         قم بالتعرف على الاجزاء التي تحتوي على الكلمات المفتاحية في الفقرة
         قم بإنشاء لغز لكل الكلمات المفتاحية بإستخدام المعلومات في الفقرة
         تأكد من انه لا يوجد اي كلمات مساعدة للوصول إلى الكلمة المفتاحية لهذا اللغز في اللغز الذي تم إنشاءه
         قم بإنشاء اللغز بحيث يدل فقط على الكلمة المفتاحية و لا يتواجد في اللغز نفسه
         تأكد من ان اللغز اجابته كلمة مفتاحية واحده فقط
         تأكد من ان لكل من الكلمات المفتاحية يوجد له لغز اذا وجد لغز مناسب\\

         مثال للمطلوب:\\
         - الفقرة كالأتي
         الفقرة: أسد حيوان من الثدييات من فصيلة السنوريات وأحد السنوريات الأربعة الكبيرة المنتمية لجنس النمور. وهو يُعد ثاني أكبر السنوريات في العالم بعد الببر، حيث تفوق كتلة الذكور الكبيرة منه 250 كيلوغراما (550 رطلًا). تعيش معظم الأسود البرية المتبقية اليوم في إفريقيا جنوب الصحراء الكبرى، ولا تزال جمهرة واحدة صغيرة مهددة بالانقراض تعيش في آسيا بولاية غوجرات في شمال غربي الهند. كان موطن الأسود شاسعًا جدًا في السابق، حيث كانت تتواجد في شمال إفريقيا، الشرق الأوسط، وآسيا الغربية، حيث انقرضت منذ بضعة قرون فقط. وحتى بداية العصر الحديث (الهولوسين، منذ حوالي 10,000 سنة)، كانت الأسود تُعتبر أكثر ثدييات اليابسة الكبرى انتشارا بعد الإنسان، حيث كانت توجد في معظم أنحاء إفريقيا، الكثير من أنحاء أوراسيا من أوروبا الغربية وصولا إلى الهند، وفي الأمريكيتين، من يوكون حتى البيرو.\\
         
         - الكلمات المفتاحية كالأتي
         الكلمات المفتاحية: أسد, حيوان, ثدييات, سنوريات, سنوريات الأربعة, جنس النمور, ببر, الزكور الكبيرة, إفريقيا, صحراء الكبرى, أمريكيتين\\
        
        هذه النتيجة: \\

        الكلمة المفتاحية: أسد
        اللغز: حيوان ثديي من السنوريات\\

        الكلمة المفتاحية: حيوان
        اللغز: ينتمي لفصيلة السنوريات\\

        الكلمة المفتاحية: ثدييات
        اللغز: نوع من الحيوانات\\

        الكلمة المفتاحية: سنوريات
        اللغز: تشمل الأسد\\

        الكلمة المفتاحية: سنوريات الأربعة
        اللغز: مجموعة من السنوريات الكبيرة\\

        الكلمة المفتاحية: جنس النمور
        اللغز: يعتبر الأسد منه\\

        الكلمة المفتاحية: ببر
        اللغز: السنورية الأكبر في العالم\\

        الكلمة المفتاحية: الزكور الكبيرة
        اللغز: تتجاوز وزنها 250 كيلوغرام\\

        الكلمة المفتاحية: إفريقيا
        اللغز: مكان عيش معظم الأسود البرية\\

        الكلمة المفتاحية: صحراء الكبرى
        اللغز: تقع إلى جنوب إفريقيا\\

        الكلمة المفتاحية: أمريكيتين
        اللغز: تتواجد الأسود فيهما\\

        شكل النتيجة النهائية :\\
        
        -اللغز: -اللغز\\
        
        -الكلمة المفتاحية: -الكلمة المفتاحية\\
\end{arabtext}}

\textbf{Arabic Prompt for Hallucination Verification: }\\
\newline
{\small
\begin{arabtext}

        قم بتقييم جودة الألغاز على حسب الفقرة الأتية\\

        الفقرة: الفقرة\\

        الألغاز: اللغز\\

        لتقوم بهذه المهمة قم بالأتي:\\
        
        قم بفحص اللغز في الفقرة. إذا كانت الفقرة تحتوي على كل من الألغاز. قم بطباعة صحيح و إذا لم تجده قم بطباعة خطأ
        تأكد من القيام بالسابق لكل لغز منفرد و طباعة النتيجة لكل لغز\\
        قم بالتعامل مع كل لغز على حدى\\
        
        استخدم الصيغة الأتية للنتيجة النهائية فقط قم بطباعة \\"اللغز:النتيجة" بدون اي شرح او اي شئ اخر
        
        الصيغة النهائية:\\

        اللغز: النتيجة\\
\end{arabtext}}
\newpage
\section{Appendix}
\label{sec:appendix_c}
In the upcoming section, you will find English translations of the Arabic content within this paper. These translations have been included to improve understanding for readers who may have limited proficiency in Arabic, ultimately ensuring greater accessibility to the content.
The translation for the Figure \ref{fig:3} content is as follows:

\textbf{Input paragraph:}

\begin{arabtext}
\small
    الذرة هي أصغر حجر بناءٍ أو أصغر جزء من العنصر الكيميائي يمكن الوصول إليه والذي يحتفظ بالخصائص الكيميائية لذلك العنصر. يرجع أصل الكلمة الإنجليزية إلى الكلمة الإغريقية أتوموس، والتي تعني غير القابل للانقسام؛ إذ كان يعتقد أنه ليس ثمة ما هو أصغر من الذرة. تتكون الذرة من سحابة من الشحنات السالبة (الإلكترونات) التي تدور حول نواة موجبة الشحنة صغيرة جدًا في المركز، وتتكون النواة من بروتونات موجبة الشحنة، ونيوترونات متعادلة، وتعدّ الذرة هي أصغر جزء من العنصر يمكن أن يتميز به عن بقية العناصر؛ إذ كلما غصنا أكثر في المادة لنلاقي البنى الأصغر لن يعود هناك فرق بين عنصر وآخر. فمثلاً، لا فرق بين بروتون في ذرة حديد وبروتون آخر في ذرة يورانيوم مثلًا، أو ذرة أي عنصرٍ آخر. الذرة، بما تحمله من خصائص؛ عدد بروتوناتها، كتلتها، توزيعها الإلكتروني... تصنع الفروقات بين العناصر المختلفة، وبين الصور المختلفة للعنصر نفسه (المسماة بالنظائر)، وحتى بين كَون هذا العنصر قادرًا على خوض تفاعل كيميائي ما أم لا.
\end{arabtext}

\textbf{English translation of the input paragraph:}

The atom is the smallest building block or the smallest part of an element that can be reached and retains the chemical properties of that element. The English word 'atom' is derived from the Greek word 'atomos,' which means indivisible, as it was believed that there was nothing smaller than the atom. An atom consists of a cloud of negative charges (electrons) orbiting around a very small positively charged nucleus at the center. The nucleus is composed of positively charged protons and neutral neutrons. The atom is the smallest unit of an element that can be distinguished from the rest of the elements. As we delve deeper into matter, we find smaller structures, but there is no longer a distinction between one element and another. For example, there is no difference between a proton in an iron atom and another proton in a uranium atom, or in the atom of any other element. Atoms, with their properties, such as the number of protons, mass, and electronic distribution, create differences between different elements, between different forms of the same element (called isotopes), and even whether this element is capable of undergoing a chemical reaction or not.

\newpage

\textbf{Extracted keywords (English and Arabic):}

Atom - {\small\<ذرة>}

Chemical Element - {\small\<العنصر الكيميائي>}

Chemical Properties - {\small\<الخصائص الكيميائية>}

Electrons - {\small\<الإلكترونات>}

Nucleus - {\small\<النواة>}

Protons - {\small\<البروتونات>}

Neutrons - {\small\<النيوترونات>}

Elements - {\small\<العناصر>}

Isotopes - {\small\<النظائر>}

Chemical Reaction - {\small\<تفاعل كيميائي>}

\textbf{Generated Clue-Answer Pairs (English and Arabic):}

\begin{arabtext}
\small
الذرة: أصغر جزء من العنصر الكيميائي يمكن الوصول إليه
\end{arabtext}
Atom: The smallest unit of a chemical element that can be reached.

\begin{arabtext}
\small
العنصر الكيميائي: يتكون من الذرات ويحتفظ بالخصائص الكيميائية
\end{arabtext}
Chemical Element: Composed of atoms and retains chemical properties.

\begin{arabtext}
\small
الخصائص الكيميائية: يحتفظ بها العنصر الكيميائي
\end{arabtext}
Chemical Properties: Characteristics retained by the chemical element.
\\

\begin{arabtext}
\small
الإلكترونات: تدور حول النواة في الذرة
\end{arabtext}
Electrons: Orbits around the nucleus in the atom.

\begin{arabtext}
\small
 النواة :  تدور حول النواة في الذرةتتكون من البروتونات والنيوترونات في الذرة
\end{arabtext}
Nucleus: Composed of protons and neutrons in the atom.
\\

\begin{arabtext}
\small
 البروتونات: تتواجد في النواة وتحمل شحنة موجبة
\end{arabtext}
Protons: Located in the nucleus and carry a positive charge.
\\

\begin{arabtext}
\small
النيوترونات: تتواجد في النواة ولا تحمل شحنة
\end{arabtext}
Neutrons: Located in the nucleus and carry no charge.
\\

\begin{arabtext}
\small
العناصر: تتكون من الذرات وتختلف بحسب عدد البروتونات في النواة
\end{arabtext}
Elements: Composed of atoms and vary based on the number of protons in the nucleus.
\\

\begin{arabtext}
\small
النظائر: صور مختلفة للعنصر نفسه
\end{arabtext}
Isotopes: Different forms of the same element.
\\

\begin{arabtext}
\smallتفاعل كيميائي: يمكن أن يخوضه العنصر بحسب خصائصه الكيميائية
\end{arabtext}
Chemical Reaction: Can be undergone by the element based on its chemical properties.
\\

Next, we will include the English translations for the examples we presented as both acceptable and unacceptable generated clues from the fine-tuned models in Tables \ref{tab:valexample1} and \ref{tab:valexample2}.
\\

\textbf{Acceptable Examples(Arabic and English):}
\\

\begin{arabtext}
\small
نجوم : في السماء ليلا
\end{arabtext}
Stars: in the sky at night  
\\

\begin{arabtext}
\small
كروم : من المعادن
\end{arabtext}
Ores: from minerals
\\

\begin{arabtext}
\small
قوة: قدرة 
\end{arabtext}
Strength: capability  
\\

\textbf{Unacceptable Examples(Arabic and English):}
\\

\begin{arabtext}
\small
زرافة:  من الحشرات
\end{arabtext}
Giraffe: from the insects 
\\

\begin{arabtext}
\small
مثلث :  مثنى مثلث 
\end{arabtext}
Triangle: plural triangle
\\

\begin{arabtext}
\small
عمة : اخت والد او والدة
\end{arabtext}
Aunt: sister of a parent or a parent's sister
\\

\end{document}